# Automotive Parts Assessment:

# Applying Real-time Instance-Segmentation Models to Identify Vehicle Parts

Syed Adnan Yusuf, Abdulmalik Ali Aldawsari, Riad Souissi
Research Department, Elm, Riyadh, Saudi Arabia

**Abstract**
The problem of automated car damage assessment presents a major challenge in the auto repair and damage assessment industry. The domain has several application areas ranging from car assessment companies such as car rentals and body shops to accidental damage assessment for car insurance companies. In vehicle assessment, the damage can take any form including scratches, minor and major dents to missing parts. More often, the assessment area has a significant level of noise such as dirt, grease, oil or rush that makes an accurate identification challenging. Moreover, the identification of a particular part is the first step in the repair industry to have an accurate labour and part assessment where the presence of different car models, shapes and sizes makes the task even more challenging for a machine-learning model to perform well. To address these challenges, this research explores and applies various instance segmentation methodologies to evaluate the best performing models.

The scope of this work focusses on two genres of real-time instance segmentation models due to their industrial significance, namely SipMask and Yolact. These methodologies are evaluated against a previously reported car parts dataset (DSMLR) and an internally curated dataset extracted from local car repair workshops. The Yolact-based part localization and segmentation method performed well when compared to other real-time instance mechanisms with a mAP of 66.5. For the workshop repair dataset, SipMask++ reported better accuracies for object detection with a mAP of 57.0 with outcomes for $AP^{IoU=.50}$ and $AP^{IoU=.75}$ reporting 72.0 and 67.0 respectively while Yolact was found to be a better performer for $AP^S$ with 44.0 and 2.6 for object detection and segmentation categories respectively.

## 1. Introduction

Automotive parts assessment is a crucial area that primarily originates from the insurance industry. The area has several other industrial applications that are gaining rapid advancement including assembly line analysis (engine parts defect detection) [1], [2], surgical instrument localization [3], healthcare (body pose, skeletal, organ, cancer/tumour segmentation)[4]–[7], botanical segmentation (plant and weeds detection/smart farming) [8]–[13], geological map analysis[14], [15], and 3D object segmentation with primary applications in autonomous/robotics point-cloud analysis and shape understanding domains [16]–[23] and construction health monitoring and assessment [24]–[26]. Most of these applications increasingly focus on real-time pixel-level assessment of area of interest with minimum impact on the overall accuracy. For certain applications such as surgery or weed removal, precision is considered paramount.

The scope of work reported in this research focusses on the automotive parts assessment and identification due to its importance in fields including accidental damage assessment, insurance claims processing, car condition assessment for rentals and other automotive portals and the automation of car repair and body shop garages. Conventional vehicle assessment processes for accidental damage repair originate from vehicle owners or enforcement authorities where a complex fault and claim assessment procedure is followed to compensate the party that is not at fault [27]. The same applies for car rental returns where drivers and rental firms come into conflict on ownership of damages to the vehicle at the time return [28]. The compensation procedures involved in these processes are often time-consuming and take weeks to complete. Moreover, a substantial proportion of the claim amount is wasted due to incorrect or unfair assessments that eventually affect the premium of the parties involved [29]. First responders and authorities also rely on visual evaluations and record keeping which still takes a long period. A real-time, and automated visual accidental damage assessment system hence offers a promising area of research.

The recent advancements made in the domain of edge and mobile device-based machine learning algorithms have made it easier to perform real-time inference of complex, vision-based tasks. MobileNets is one such, lightweight deep neural network, specifically focussed on hyperparameter tuning to maintain a trade-off between latency and accuracy of the classifier [30]. Similarly, DeepDecision is another architecture that focusses on balancing trade-offs such as impact of video resolution, parameterizing accuracy, end-to-end latency, and video compression to develop an edge video analytics deep learning framework [31]. On the object bounding box estimation mechanisms RCNN [32], Fast RCNN [33], Faster RCNN [34], SSD [35], and YOLO [36], are a few well-known algorithms. In industrial applications, faster detection and segmentation mechanisms are increasingly used in domains such as defect/weld image surface inspections [37]–[40], manufacturing/textile lines quality analysis [41], smart crops and farming [42], semiconductor fabrication and design processes [43] and engineering condition monitoring fault diagnosis [44]–[46] due to their faster and more accurate performance.

These frameworks are known to require a large amount of training data. Moreover, for the current scope of work, identification of automotive parts cannot only be limited to mere bounding boxes due to complexity of automotive part and damage outlines and hence it is difficult to get an end-to-end detection. Moreover, automotive datasets are extremely large due to the variability of model variants and makes. Hence, at the feature extraction stage, an increase in convolutional layers lead to gradient disappearance or explosion. To address these challenges, an array of deep network architectures has been introduced with their own strengths and weaknesses. He Kaiming *et al.* proposed a residual network (ResNet) that assists in model convergence by utilizing a residual module accelerating the neural network training process [47] by combining it with the target detection model via Mask R-CNN [48]. The Mask R-CNN algorithm is among the first deep learning mechanisms that combine the object detection and segmentation techniques improving the overall identification accuracy. Other similar deep learning variants in this domain include the pioneering AlexNet [49] to VGGNet [50], GoogLeNet [51], and ResNet [47], [52].

A major drawback in these architectures have been their primary focus on achieving higher accuracies instead of improving latency and performance aspects which makes them unsuitable for real-time image processing. Within the scope of car parts segmentation, most of the object detection and segmentation techniques offer two major shortcomings, (1) the ability to process images in real time and (2) the model's ability to segment adjacent parts uniquely. Once such case is that of front and rear doors which, in many cases, have very little separation to be identified uniquely. This context leads to a well-known image process principle known as "instance segmentation". The domain is one of the most performance-intensive methodologies in deep-learning since several unique classes are often adjacent and have high similarity. So far, previous models have aimed for accuracy over speed.

For mobile/edge applications, smaller network architectures are more efficient in distributed training, require less bandwidth during remote model updates and, most importantly, feasible to deploy on smaller, mobile and edge-devices [53]. Moreover, limited space on mobile and edge devices also remains a major challenge. Regardless of large storage sizes, installing a mobile application that requires pulling 500MB of memory in the form of a weights file is a substantial discouraging factor for users pertaining to the fact that most smartphone applications do not go above 100MB in size. Model pruning is a well-known that utilized to make smaller and more efficient neural networks. The technique involves eliminating unnecessary values in the weight tensors of a neural network resulting in a compressed network that runs faster and has a reduced computational cost during network training [54].

This research explores on the context of utilizing and improving on the existing mobile/edge-device level deep neural network algorithms for the purpose of car part segmentation. The methodology employs various object-detection architectures to identify vehicle parts. The phase is crucial in a sense that identification of missing vehicle parts via a standard object detection algorithm presents high true negative (missing part identification) or false positive (missing part identified in non-vehicle background) cases due to the high variability of backgrounds that are visible in missing car parts.

In the presented case, the part detection methodology trains on a total of 29 car part types with each part then further labelled to have three unique damage categories including scratches, minor and major dents. The severity of each damage type is then mapped to generate an overall damage level of the said category as a regression score. The purpose of this generalized scoring is due to the unavailability of a single parts and labour cost prediction paradigm due to a variability of these in different world economies.

The contributions of this research are:
1. Development of a comprehensive car parts dataset:
    a. The second dataset was extracted from national government databases engaged in accident information and data management activities for law enforcement, insurance, and other purposes.
    b. The third dataset involved data augmentation and class merging phases in a bid to improve on the overall segmentation and detection accuracies by minimization cross-class similarity
2. Implementation of a dual-stage, real-time instance segmentation algorithm (Yolact) [55]
3. Comparison of the two-stage detection mechanism with a single-stage instance segmentation mechanism (SipMask) and its variant (SipMask++)[56]
4. Comparison of four algorithms against:
    a. An initially extracted third-party dataset (DSMLR)
    b. Internally extracted dataset mentioned in 1

The paper is organized as follows:

Section 2 presents a review of existing deep-learning methodologies, the existing vehicle parts and damage assessment processes, and the scope of research. Section 3 presents the architecture including the instance segmentation methodologies employed including YOLACT, SipMask, SipMask++ architectures. The section presents an in-depth performance comparison of these architectures. Section 4 ultimately concludes with the key outcomes' discussion and future directions of this research.

2. **Background & Overview**

Industrial visual inspection is an area gaining rapid advancement and interest. Advanced computer vision and deep-learning methodologies are explored to facilitate automation while addressing problems such as weakly annotated/sparse datasets [57], depth-wise separable convolutions (MYOLOV3-Tiny) [58], mixed-supervision annotation for quicker and better identification model training [59], overlapping or complex region localization [60]. Many of these techniques focus on the improvement of the deep-learning modelling pipeline either by diversifying the data extraction process, tuning the pipeline parameters, enhancing the training mechanism, or evolving the existing architecture to gain on the underlying segmentation or detection techniques.

The scope of research presented in this work focusses on the evaluation and application of real-time segmentation methods on the identification of complex part groups in vehicles. The layout of automotive parts follows several spatial constraints that can be very strict for cases such as sidemirrors connected to doors or diverse such as backdoor window enclosed within the rear door or in the body of the vehicle itself (for two-door vehicles). These constraints can only be addressed via statistical multi-model distribution models. Active Appearance Models [61] and Deformable Part Models [62] are two such models that are used to ascertain, in advance, various combinations parts may form in a vehicle. Such combinations may not be sufficient to cover all the shape groups due to the variations induced in automotive images due to angle, rotation, and object deformation.

Deep Convolutional Neural Networks have recently shown outstanding results in a range of computer vision problems. During the past decade, a lot of interest and research has been focussed on aspects such as hyperparameter tuning and optimization, network pruning and connectivity learning to improve on the model size and performance specifically on mobile and edge devices [63]–[65]. Moreover, the focus is also on model pruning [66]–[70] and connectivity learning to gain on model size and network performance [71], [72]. Semantic and instance segmentation mechanisms are two object shape estimation techniques that are commonly used to identify object boundaries.

At this stage, a detailed information extraction regarding the vehicle is performed. In conventional cases, this may include marking a vehicle template to record damage areas that are specific to the accident and can also include the severity of the damage as well. The party involved in the accident or the officer in charge of the record-keeping may also take images and videos for insurance and other investigation purposes. The information is then consolidated by the relevant authorities, such as the insurance agencies, to prepare damages grant claim to be given to the affected parties.

The objective of a smart damage assessment would be to automate and streamline the entire process. In most vehicular damage cases, the assessment may comprise of three core phases (Figure 1):

1. **Evidence data extraction**: Extraction of visual evidence such as pictures and videos taken to clearly record the damage. It is ensured that the damage context is suitably preserved such as the images are taken from a suitable distance to allow the algorithm to differentiate between various car parts involved.
2. **Automated parts identification**: The visual information is then submitted to a vehicle part identification algorithm which utilises pre-trained AI models to identify the boundary part. This phase includes the vehicular brand, type, and age of the vehicle to select the identification model relevant to that model type. For instance, a model trained to predict a 3-year-old sedan cannot be used to identify the bumper of a less-then-a-year-old SUV.
3. **Automated damage assessment**: This phase includes the damage area extraction of each car part along with the AI logic that estimates the part cost as well as the labour estimate.
4. **Damage recommendation output**: The part and damage information are then modelled against other relevant information such as the car model and type to generate an accurate damage report. This report may contain assessments ranging from car part price to the overall labour cost the repair will need. The report may also generate a recommendation if a complete replacement of the part is needed.

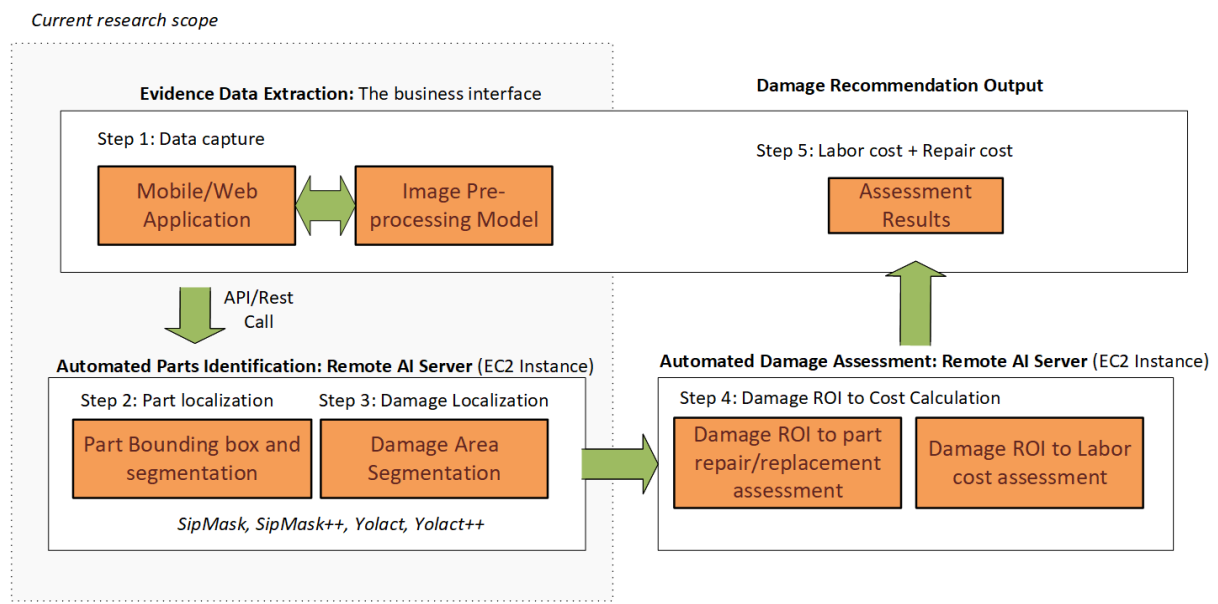

*Figure 1: Description of scope research presented as well as the overall car damage assessment use case*

## 2.1. Anatomy of car part damage:

A vehicle comprises of a set of parts that are joined together in a complex assembly where the distribution of damage cannot directly be ascertained. This means that damage done to a front bumper will have a different cost assessment paradigm than a door. Moreover, different severity levels of parts will also have a variable impact. For instance, back bumper and chassis damage to a rear-engine vehicle would have a different assessment regime for a major damage then a front-engine vehicle. In a conventional setting, a damaged car is first assessed by a designated workshop where the outer damage is first categorized to belong to one of many damage types. Based on the severity of damage, a decision to either part repair, replacement or a complete writing-off of the vehicle is made. The part repair cost estimate is often diverse for cases such as scratches and minor dents, depending upon the spread of damage to each part as well as the part type. For instance, a minor, scratch-less dent would often need a pull-correction without repainting Figure 2(a) whereas a scratch and dent combination on a vehicle door would not only need a larger surface area to be repainted but the structural deformation to be correct as well Figure 2(b). On the other hand, a damage leading to a substantial structural deformation may generate a total replacement pay out (vehicle written-off). Minor dents are often corrected via special part-moulds and heat guns whereas scratches are repainted to the part colour code.

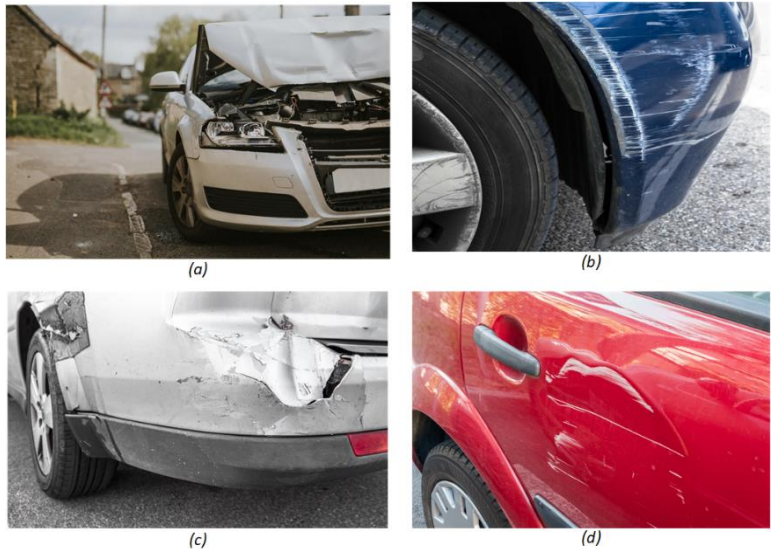

*Figure 2: Various damage categories with variable repair outcomes including (a) deformed/missing parts, (b) scratches (c) major dent and (d) scratch and dent combination*

### 2.2. Identification of missing parts:

Another major challenge in car part identification is a missing part itself as is quite common in detachable vehicle sections such as bumpers, side mirrors and wheel caps Figure 2(d). Due to a high variability of backgrounds in missing car parts, it is often challenging to classify a class absence merely via the background of that part. This is a very common case in major accidents where a car part is either completely absent or distorted to the extent that it cannot be visibly identified. One such example is that of the left fender shown in Figure 2(c).

### 2.3. Generalizing model for accurate segmentation:

Another major challenge reported later in this research is that of training a single, generic car part identification model as presented by Pasupa *et al*.[73] where well established deep learning models such as Mask R-CNN and GCNet generated low mAP outcomes ranging around 48.5 for GCNet to 54.3 for HTC over ResNet-101 encoder for part detection and 43.0 to 65.2 for GCNet and CBNet for Resnet-101 and ResNet-50 respectively for part segmentation. The lower rates for both part localization and segmentation form a substantial basis of this research as this approach utilised a single model to identify car parts identification for a wide range of designs despite the high variability of the automotive designs.

### 3. Proposed System Architecture and Methods

Most reviews in vehicle damage assessment so far have been focussed on two core areas of car parts segmentation and damage assessment. Both the stages are generally connected sequentially to have a more accurate assessment of the part damage based on the model, year, and actual price of the car at the time of evaluation. The damage assessment phase is then further divided into a parts cost and labour cost prediction models in addition to advisory aspects that vary from case to case depending upon the fact if the underlying policy requires replacement based on the level of damage incurred to that part.

A model trained for a specific group of cars (e.g., sedans or SUVs) may be reasonably accurate for certain parts but a single, universal damage assessment model cannot be efficiently trained. This research makes several contributions to vehicle parts segmentation. Firstly, it compares the performance of four deep learning models by Pasupa *et al*. namely Mask R-CNN, GCNet, PANet, CBNet and HTC over the same dataset on Yolact, Yolact++, SipMask and SipMask++ algorithms. It then presents a more detailed part detection architecture on a dataset collected in-house and compares the performance of the two dual-stage instance segmentation mechanisms (Yolact) against two single-stage mechanisms (SipMask/SipMask++).

### 3.1. Deep-learning methodologies for car parts segmentation

Car parts identification can be addressed from two unique contexts in deep learning. One common approach originates from part detection where each part is labelled as a bounding box. However, since a car is a symmetrically complex combination of smaller parts/components, annotating each part with a matching background creates a lower cross-class variability which may lead to a higher level of cross-class mismatches. The other method is to have a pixel-level segmentation of each of the car parts leading to a more accurate, polygonal representation of each of the car parts. This technique is considered more reliable though the inference time for a semantic or instance segmentation mechanism has so far been a computationally expensive task which has focussed more on accuracy then performance so far. Recently, real-time object segmentation has gained traction due to improved hardware and methodological improvements in soft computing techniques [74]–[76]. In deep-learning, various deep-learning architectures managing framerates of 40+ fps for the instance segmentation. Mask RCNN is the most common instance segmentation technique that comprise of a 2-stage modelling mechanism that involves an object proposal stage extending to a segmentation calculation, mask, class confidence and bounding box offset estimation stages. Hence, the second stage is essentially the calculation of per-instance mask coefficients. Since the two tasks run in parallel, the segmentation process is substantially faster than the other methods. The parts detection system was evaluated on three major variants of instance segmentation techniques including Yolact, SipMask, and SipMask++.

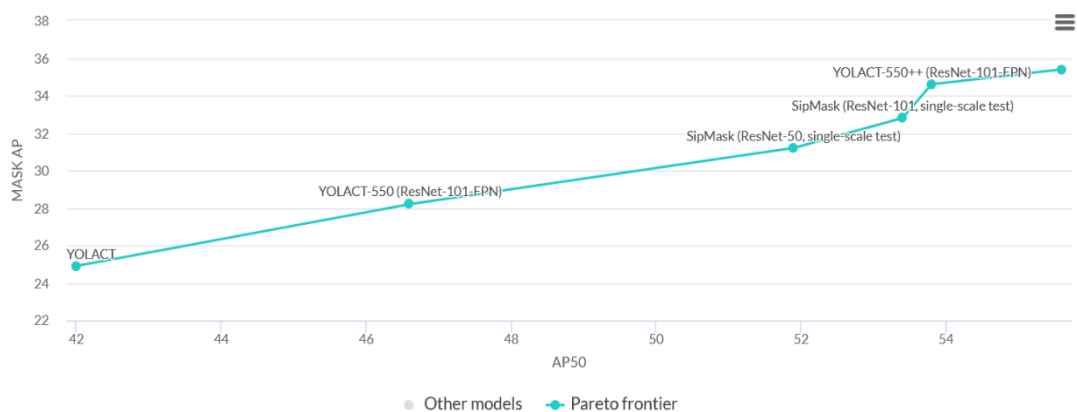

*Figure 3: A performance comparison of real-time instance segmentation techniques including YOLACT [55]/YOLACT-550 (ResNet-101), SipMask (ResNet-50)/SipMask (ResNet-101)[56] and Yolact-550++ (ResNet-101-FPN)[77] on the COCO dataset*

Yolact is a recent proposed instance segmentation mechanism reported to have a superior trade-off between performance and accuracy by predicting a dictionary of basis masks (category-independent maps) for an image and a set of coefficients that are instance-specific. Yet, the method has inferior results reported when compared to two-stage methods. Yolact is an instance segmentation mechanism presented by the *Bolya et al.* as the first method attempting real-time instance segmentation [55]. Since within the scope of this research, instances play a significant role in car parts segmentation, this research primarily focusses on the evaluation of Yolact [55] and extends and compares its performance against other, well-known and reporting segmentation mechanisms including SipMask, SipMask++ and Yolact++ paradigm [48], [56], [77], [78] Figure 3. So far, SipMask++ is known to have reported the best Mask AP (35.4) followed by YOLACT-550++ (34.6) on a ResNet-101 backbone. The framerate, however, is where YOLACT precedes by 45.3 fps to SipMask 41.7 and SipMask++ at a significantly lower rate of 27 fps. It is the outstanding performance of the YOLACT and SipMask regimes that inspires the evaluation of these techniques for the domain of car parts segmentation.

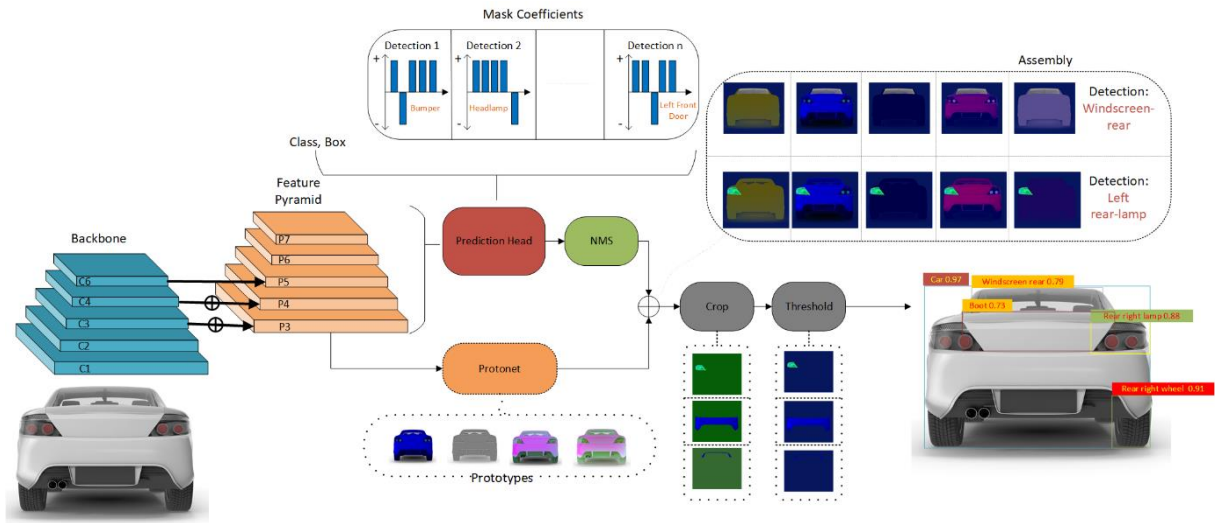

*Figure 4: YOLACT architecture explained for the car parts segmentation use case*

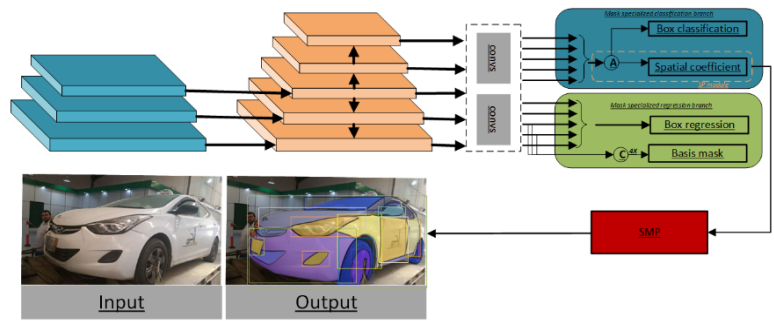

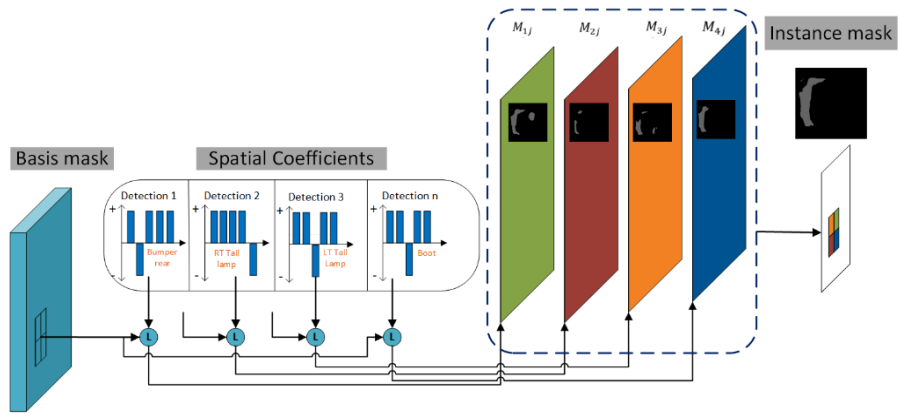

*Figure 5: The SipMask architecture within the car parts segmentation context covering (a) the feature alignment and spatial coefficient generation process and (b) the basis mask to spatial coefficient generation process leading the final instance segmentation of the 'Bumper-rear' instance.*

The SipMask module is a lightweight spatial preservation technique that preserves the spatial information for each car part within a bounding box. The approach is a single-stage instance segmentation method aimed at faster inference speeds by avoiding proposal generation and feature pooling stages. Yet, the accuracies of single-stage-approaches have poor results. Figure 5 shows the overall architecture of the SipMask instance segmentation mechanism used for car parts segmentation and comprises of fully convolutional mask-specialised classification and regression branches. The SipMask design focusses on the spatial preservation (SP) module in the mask-specialised regression branch and performs the tasks to align features and generating spatial coefficients. In this approach, each predicted bounding box has a separate set of spatial coefficients. These spatial coefficients preserve the spatial information contained within each object instance and hence allows a better definition of adjacent spatial objects especially where the spatially similarity of such objects is high. One such example is that of the front and rear car doors where the only separation between the two is often a vague line. The mask-specialized regression branch, on the other hand, predicts the bounding box offsets as well as a set of basis masks that are category independent. Figure 5 shows an example car image where a set of spatial coefficients for a basis mask are generated that include bumper, rear lights, boot, etc. The SipMask case and Yolact comparison is shown in Figure 6 where only the case of 'bumper' instance is considered for the sake of simplicity. The comparison of corresponding mask generations for both Yolact and SipMask is shown in (a) and (b). The map $M_j$ shows a linear combination of a set of single set of coefficients and basis masks. In Yolact, the final mask $\widehat{M}_j$ is obtained by pruning and thresholding the mask $M_j$. Figure 6 (b) shows the second (k=2) quadrant spatial coefficient generation for a bounding box $j$. This results in a separate set of spatial maps $M_{ij}$ where $i$ is the quadrant number for the bounding box $j$. The spatial maps are then pruned and integrated via a simple addition and thresholding to obtain the final map $\widehat{M}_j$. This spatial relationship reduces the influence of the adjacent 'boot' instance and generates a better mask prediction. The process is further repeated for a better spatial comparison of 'boot' to 'LT Tail lamp' and 'RT Tail lamp' instances.

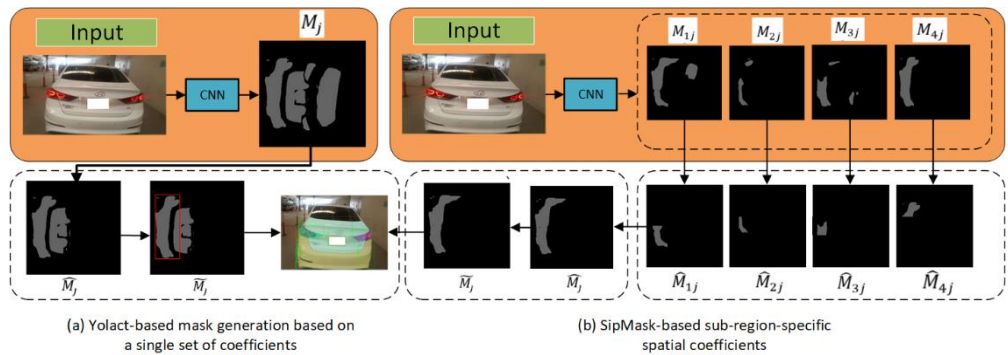

Figure 6: A Yolact to SipMask comparison with (a) A single coefficient set (non-spatial) example compared against (b) sub-region-specific spatial coefficient generation via SipMask

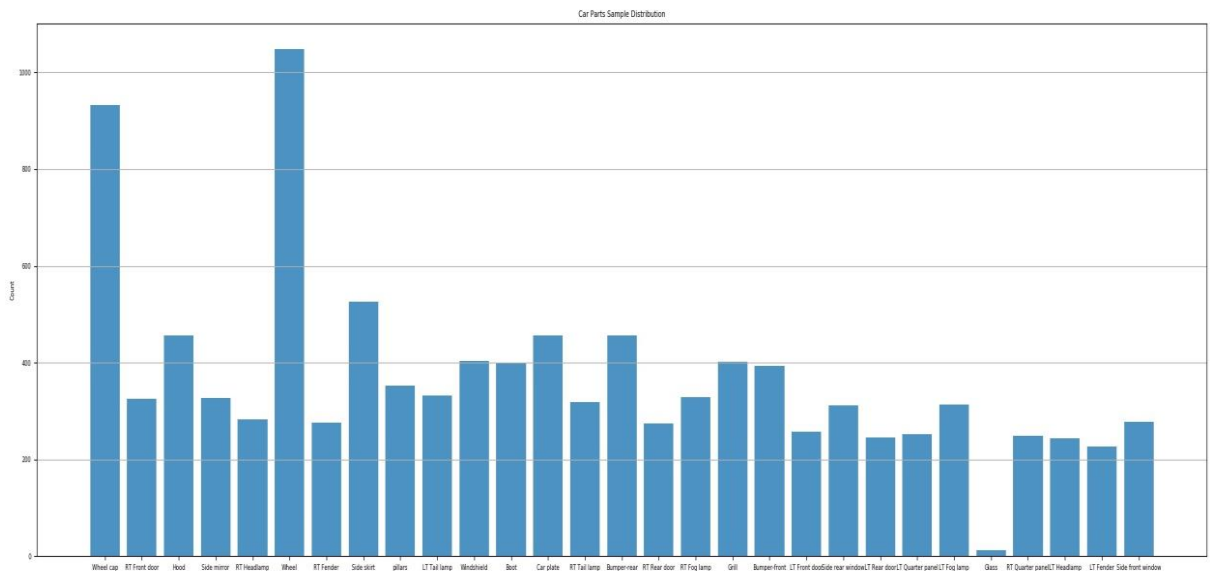

Figure 7: Distribution of various car part types in the dataset including 29 different classes

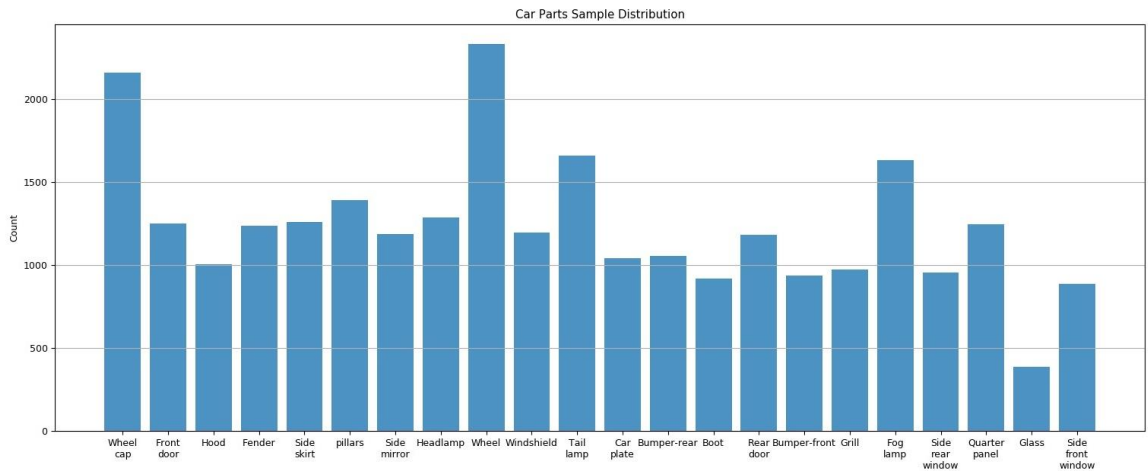

*Figure 8: Distribution of various car part types merged in the dataset including 22 different classes*

### 3.2. Dataset Description

In the research community, at present, there is only a limited number of datasets available in public. Pasupa *et al.* [73] have the most recent car parts dataset comprising of 500 car images taken from various angles for a wide range of cars with 18 car parts masks and bounding boxes. This dataset was used to train a Yolact model mainly to assess the suitability of the dataset to segment car parts. Due to high variability of the dataset covering sparsely distributed class groups and models, the data generated a very low BBox mAP of 0.33 improving to MAP50 accuracy of 0.65 only. This deemed it necessary to have a more organized and balanced dataset aimed at specific car types. Since the aim of this research was to perform damage cost assessment, car model also played a substantial role. Hence, training a part detection model from a variable range of car models would also have generated parts and labour cost inaccuracies at later stages. Hence, within the scope of this research, the car model considered for data extraction was Hyundai Elantra models from the past 7 years.

The dataset for this research included a total of 1032 images containing 11707 annotated car parts originated from 29 unique classes (Figure 7). Most images were extracted from pictures taken from car workshops where the vehicles were brought for repairs. The initial extraction process was arbitrary, and the pictures were taken for the shops' own record keeping purposes. The existing results are reported on these arbitrary images though the data collection method has since been organized substantially for model retraining and further tuning stages and includes the capturing of some or all instances of the car. The data did have an inherent bias towards the 'Wheel' and 'Wheel_cap' categories (Figure 7) though due to the unique spatial characteristic of these classes, the bias was assumed to have lesser impact on any other parts.

### 3.3. Training of the Instance Segmentation Models

The training objective of this research was to learn features of parts to identify and segment these parts in unseen images. Since the goal of this research is to identify and assess damage, the presumption here was that damage done to any model types could more broadly be identified on that specific model and not on any car models in general due to a variable vehicle cost range and hence the rationale behind focussing only on a single car model. The model selection specifically focussed on the following objectives:

- The model's ability to separately identify overlapping parts
- Whether the model was capable of processing both part localization and segmentation in real time
- The precision of both the localization and segmentation routines

Based on the evaluation done in Section 3.3, two algorithm genres were evaluated along with their extensions. Among these SipMask/SipMask++ differed from the Yolact regime due to its single stage segmentation network that elevated the segmentation performance without any trade-off in the speed by a novel architecture called Spatial Preservation (SP) in which the network generated a set of spatial coefficients per box prediction. During the originally reported tests on the COCO dataset, this preserved the information of adjacent objects. As both underlying base architectures for YOLACT and SipMask are the same, the differences come to play at their heads. The two heads in YOLACT are: (1) Anchor-based regressions to produce bbox (bounding box), class, and the coefficients. (2) The Protonet, to produce image sized masks. Both the branches are combined into a single network at the end to produce a rectified mask along with the bounding box. In contrast in SipMask, single regressor yielding bases mask and bounding box that are fed thereafter to the ConvNet for generated spatial coefficient and box classification. the novelty in SP module where the coefficients and basis masks divided into K x K region to preserve and delineate the adjacent objects masks from each other. SipMask++ claims of an exceptional performance on a level of small pictures (~550 x 550), while the original Yolact outperformed others for big pictures (~1330 x 800). We also observed better performance of classification in two stages networks like Yolact. Our reasoning is that separate network dedicated for such task should present improved segmentation results. Moreover, there are plans in future to integrate DarkNet as a backbone since it has generated good performance in terms of bbox and classification, which is likely to play a substantial role in the optimization of the network as well as in the incorporation of the Bayesian layer or graph similarity nodes to get adjacency information benefits.

### 3.4. Experimental setup and results

To train the ML model for car parts segmentation, the system was trained on three car parts datasets as follows:

#### 3.4.1. Dataset 1 (DSMLR):

Initial DSMLR dataset with 18 classes with instance masks and bounding boxes
- Dataset size: 500 images of sedans, pickups, and SUVs predominantly scraped from online images
- Format: The dataset is available in COCO Challenge format from https://github.com/dsmlr/Car-Parts-Segmentation [73].
- Preparation: The images were normalised into 1024 x 1024 pixels with zero padding used to maintain the aspect ratio. The dataset was randomly partitioned into a training set (70%) and a test set (30%). The model was trained to 300 epochs while saving the best model for the epoch generating the lowest validation loss.
- The losses were calculated for
    o Classification loss,
    o Localization loss, and
    o Parts segmentation task loss.
- A Cross Entropy loss was used to calculate validation losses with the Stochastic Gradient Descent (SGD) method for parameter optimization with a learning rate of 0.1 and weight decay of 0.0015. The experiments were run for the pre-determined epochs and the best intermediate model was stored despite the training running up until the entire number of epochs.

#### 3.4.2. Dataset 2 (3.2Workshop repair dataset):

The workshop repair dataset gathered as described in Section 3.2 with 29 classes with instance masks and bounding boxes:
- Dataset size: 1032 images of sedans, pickups, and SUVs
- Preparation: The images were normalised into 550 x 550 pixels with the dataset randomly partitioned into a training set (70%) and a test set (30%). The model was trained to 300 epochs while saving the best model for the epoch generating the lowest validation loss.
- The losses were calculated for
    o Classification loss,
    o Localization loss, and
    o Parts segmentation task loss.
- A Cross Entropy loss was used to calculate validation losses with the Stochastic Gradient Descent (SGD) method for parameter optimization with a learning rate of 0.1 and weight decay of 0.0015. The experiments were run for the pre-determined epochs and the best intermediate model was stored despite the training running up until the entire number of epochs.

### 3.4.3. Dataset 3:

For this dataset, the number of samples was increased to 2200 images while also incorporating a low number of images of different brands though belonging the same vehicle type i.e., sedans. These images represented less than 20% of the dataset. In addition, to overcome the shortage of the dataset samples, removing the positional labelling of some car parts helped improving the model performance by significant margin. The class merger reduced the total number of classes to 22, down from 29 unique classes. Positional labels meant for the classes that had left or right classes for a same class type (e.g., the side doors) a single class label was allocated.

**Data augmentation**: The initial technique explained in Figure 6 exposed the underlying models to a wide variety of test images with substantial variations in backgrounds, unique car colours, unclear, angled or body parts or flipped images. Moreover, there was substantial similarity between left and right car parts such as doors, or mirrors that resulted in a significant cross-class bias. To improve on these shortcomings, the following augmentation steps were undertaken:

- Augmentation to multiply less represented classes such as the window glass
- Addressing light intensity variations by adjusting gamma values
- Merging side-classes (e.g., Left and Right tail lamps)
- Flipping to adjust for left and right-sided images both horizontally and vertically
- Incorporating rotations since the dataset was found to be skewed at certain angles by injecting standardized images as simple FCN on top of a VGG16 backbone with four rotation categories (0, 90, 270, 360)
- To compensate for the colour variation, the model was made colour agnostic by introducing RGB shift, contrasting, and normalization
- To address the vague part boundaries, blurring and hue saturation was used

Moreover, since very small objects showed a low mAP, a copy-paste augmentation methodology was introduced [79]. Doing so improved the baseline results of SipMask substantially. The technique generated new training images with different backgrounds, part-visible car-parts, and scattered car parts (e.g., detached bumpers). Unlike other conventional datasets where objects can be isolated from the background, the bounding box regressor benefitted from this multi-part complexity of the car parts dataset.

### 3.5. Performance evaluation metrics

At the training, each of the segmentation algorithms compared the predicted polygons with the ground-truth based on the Intersection over Union by updating its parameters at each iteration as follows:

$$IoU = Area\ of\ overlap/Area\ of\ union$$

The underlying principle as per the COCO Challenge was to obtain an $IoU \geq 0.5$ hence indicating any overlap of more than or equal to 50% to be a true prediction. For the car parts prediction use case, this threshold was kept as it stood. Each of the four algorithms presented in Table 2 was evaluated for 6 out of 12 detection evaluation metrices defined by the COCO Challenge including the Average Precision primary challenge metric $AP$, PASCAL VOC metric $AP^{IoU=.50}$, and strict metric $AP^{IoU=.75}$ and Across Scales AP of $AP^{small}, AP^{medium}, and\ AP^{large}$ [80]. The $mAP$ was calculated with the average of 20 Intersection over Union (IoU) values of each object between the thresholds 0.5 and 0.95 as follows:

$$mAP = \frac{\sum_{i=0}^{14} AP_{50+(2.5*i)}}{20}$$

The $AP^{IoU=.50}$ was calculated as a single IoU of 0.50 and 0.75 corresponding to the metrics $AP^{IoU=.50}$ and $AP^{IoU=.75}$. The $AP^{IoU=.50}$ and $AP^{IoU=.75}$ metrics meant that the IoUs were greater than or equal to 0.5 and 0.75 intersections of the original and detected bounding boxes.

### 3.6. Experimental results and discussion

The section presents the performance comparison of the three algorithms (SipMask++, SipMask and Yolact) against the DSMLR, Repair Workshop datasets (both non-augmented and augmented).

The comparisons made are based on:

1. The instance/semantic segmentation results of car parts against their ground-truth
2. The robustness at various arbitrary angles and zoom levels
3. The computation efficiency (frames per second)

#### 3.6.1. Dataset 1

The section presents the performance comparison on the DSMLR dataset against the 5 algorithms (Mask R-CNN, GCNet, PANet, CBNet, HTC) reported by Pasupa et al. against the three algorithms (Yolact, SipMask, SipMask++) compared in this research. As per the Table 1, Yolact presents a marked improvement in the overall object detection accuracy with a mAP of 61.3 for object detection against the HTC accuracy of 54.1. On the segmentation performance, Yolact outperformed HTC with mAP of 66.5 while GCNet performing better on the $AP^{.50}$ with 78.2. In general, the $AP^s$ values are reported to be the worst performing accuracies where Yolact performed slightly better (42.6) compared to PANet (38.5).

| Model | Object Detection | | | | | | Instance Segmentation | | | | | |
|---|---|---|---|---|---|---|---|---|---|---|---|---|
| | $mAP$ | $AP^{.50}$ | $AP^{.75}$ | $AP^s$ | $AP^m$ | $AP^l$ | $mAP$ | $AP^{.50}$ | $AP^{.75}$ | $AP^s$ | $AP^m$ | $AP^l$ |
| SipMask++ | 51.1 | 69.2 | 58 | 33.9 | 41 | 58.4 | 55 | 77 | 59 | 32.1 | 54 | 59.2 |
| SipMask | 49.7 | 71.6 | 60.5 | 30.3 | 48.5 | 59.8 | 53.7 | 71.3 | 62.1 | 30.7 | 48.1 | 61.2 |
| Yolact | 61.3 | 61.3 | 60.8 | 39.5 | 60.5 | 63.1 | 66.5 | 67.1 | 58.8 | 42.6 | 44.5 | 65.7 |
| | Object Detection | | | | | | Semantic Segmentation | | | | | |
| Mask R-CNN | 51.1 | 69.2 | 58.0 | 33.9 | 41.0 | 58.4 | 55.0 | 77.0 | 59.0 | 32.1 | 54.0 | 59.2 |
| GCNet | 50.9 | 76.8 | 57.7 | 32.3 | 45.6 | 56.7 | 54.6 | 78.2 | 63.9 | 34.9 | 48.3 | 61.7 |
| PANet | 48.8 | 76.5 | 56.4 | 32.9 | 48.6 | 51.8 | 54.0 | 77.3 | 63.5 | 38.5 | 51.4 | 58.7 |
| CBNet | 51.9 | 71.6 | 60.8 | 28.6 | 48.3 | 57.9 | 53.0 | 72.2 | 63.0 | 28.6 | 48.3 | 61.7 |
| HTC | 54.1 | 75.7 | 63.6 | 34.4 | 50.4 | 61.1 | 55.2 | 76.1 | 65.2 | 36.1 | 50.2 | 63.6 |

*Table 1: A performance comparison ranking with the DSMLR dataset and the performance reported in Pasupa et al [73] against the algorithms reported in this research SipMask++, YOLACT-550, SipMask and YOLACT*

#### 3.6.2. Dataset 2

The performance output of Dataset 2 is shown in that includes $mAP$ and $AP$ at various thresholds with the ResNet-101 backbone. As per the table with regards to, SipMask++ showed the highest $mAP$ of 0.57 along with
the best outcomes for $AP^{IoU=.50}$ and $AP^{IoU=.75}$ of 0.72 and 0.67 respectively for object detection and 0.65 and 0.44 for instance segmentation categories. Yolact was found to be a better performer for $AP^s$ with 0.44 and 0.026 for object detection and segmentation categories respectively.

| Rank | Model | Object Detection | | | | | | Instance Segmentation | | | | | | fps |
|---|---|---|---|---|---|---|---|---|---|---|---|---|---|---|
| | | $mAP$ | $AP^{.50}$ | $AP^{.75}$ | $AP^s$ | $AP^m$ | $AP^l$ | $mAP$ | $AP^{.50}$ | $AP^{.75}$ | $AP^s$ | $AP^m$ | $AP^l$ | |
| 1 | SipMask++ | 57 | 72 | 67 | 39 | 64 | 71 | 43 | 65 | 44 | 2 | 12 | 47.2 | 17.5 |
| 3 | SipMask | 44 | 58 | 51 | 29 | 50 | 59 | 39.7 | 62.1 | 42 | 0.7 | 11.1 | 43.8 | 20.8 |
| 2 | Yolact | 49 | 63 | 53 | 44 | 55 | 66 | 41.2 | 65 | 43 | 2.6 | 10.5 | 45.7 | 21.1 |

*Table 2: A performance comparison ranking of SipMask++, SipMask and YOLACT instance segmentation performance against the workshop repair dataset with merged, 22-class Dataset 3*

Figure 8 presents a visual comparison of detection and segmentation results with each of the three algorithms (Yolact, SipMask and SipMask++) shown row wise from top to bottom. For Car A, all three algorithms failed to segment the left and rear doors. At straighter angles (Car B), the doors' classes were detected correctly but Yolact and SipMask missed the back fender detection and segmentation. In general, SipMask++ showed more instance segmentation resilience for both smaller and large parts with some masking overlaps/inaccuracy in (e) along with a low parts localization accuracy. Parts localization (detection) had best results for all three cars via SipMask (g-i). In terms of overall computational efficiency, SipMask++ again took a lead with an average of 17.5 fps followed by the second-best performer that was SipMask with 20.8 fps. A consensus on these results was that SipMask++ showed better localization/detection and an $mAP$ of 0.57. This was followed by Yolact with an $mAP$ of 0.49 and 0.41 for detection and segmentation respectively. Yolact was found to handle $AP^s$ better for both detection and segmentation.

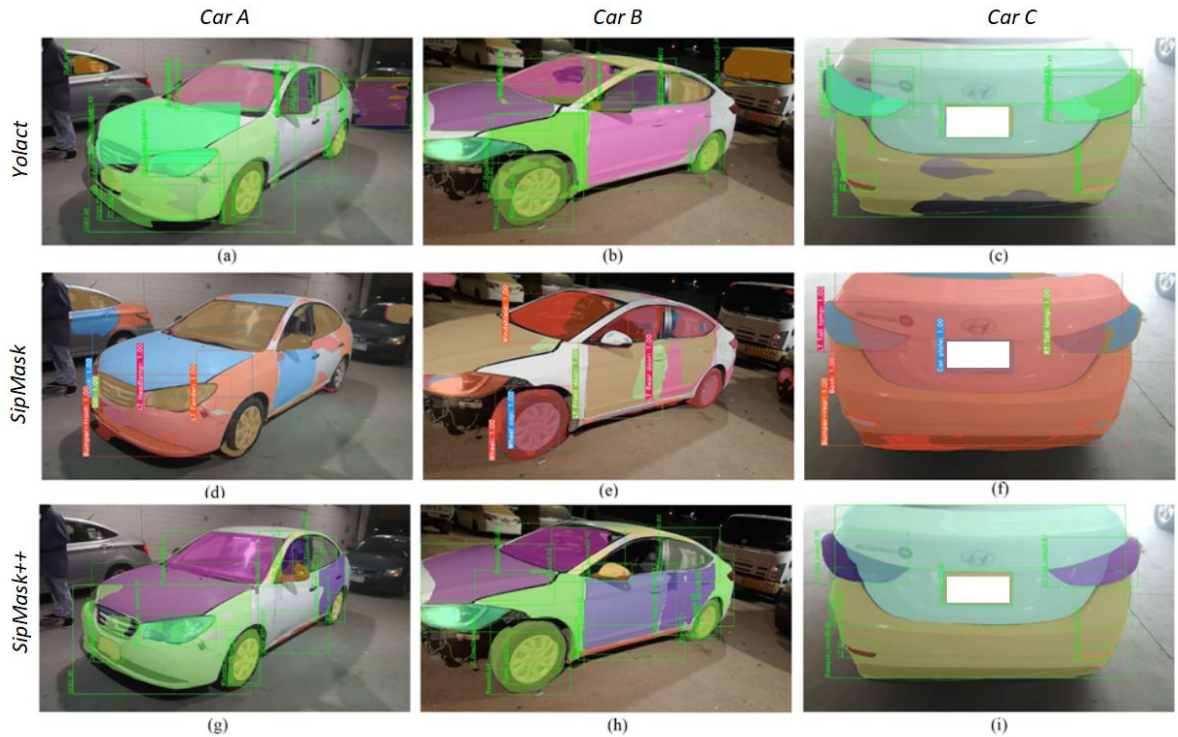

*Figure 8: Comparing Yolact to SipMask/SipMask++ instance segmentation regimes. (a-c)Yolact, (d-f)SipMask, (g-i) SipMask++*

### 3.6.3. Dataset 3

The performance on the merged classes' dataset results is shown below (
Table 3). A significant improvement can be noticed on all aspects including the bbox and mAP scores. The outcomes showed the SipMask genre to be an overall outperformer when compared against the Yolact algorithm except for the smaller bounding boxes. Yolact and SipMask adjusted far better at the introduction of augmentation and class adjustment measures.

*Table 3: A performance comparison ranking of SipMask++, YOLACT-550, SipMask and YOLACT instance segmentation performance against the workshop repair dataset*

| | | Object Detection and Instance Segmentation (DGX) | | | | | | | | | | | |
|---|---|---|---|---|---|---|---|---|---|---|---|---|---|
| | | Object Detection | | | | | | Instance Segmentation | | | | | |
| Rank | Model | $mAP$ | $AP^{.50}$ | $AP^{.75}$ | $AP^s$ | $AP^m$ | $AP^l$ | $mAP$ | $AP^{.50}$ | $AP^{.75}$ | $AP^s$ | $AP^m$ | $AP^l$ | fps |
| 2 | SipMask++ | 55.9 | 84.4 | 63 | 2.5 | 26,6 | 60.8 | 50.6 | 77.5 | 54.9 | 2 | 17.7 | 56 | |
| 1 | SipMask | 58.8 | 85.9 | 66 | 2.3 | 28.3 | 63.8 | 53.3 | 80.2 | 57.3 | 1.1 | 19.2 | 59.3 | |
| 3 | Yolact | 44.6 | 80 | 45.7 | 8.4 | 24.1 | 47.1 | 49.2 | 74.6 | 53.5 | 2.8 | 24.5 | 53 | |

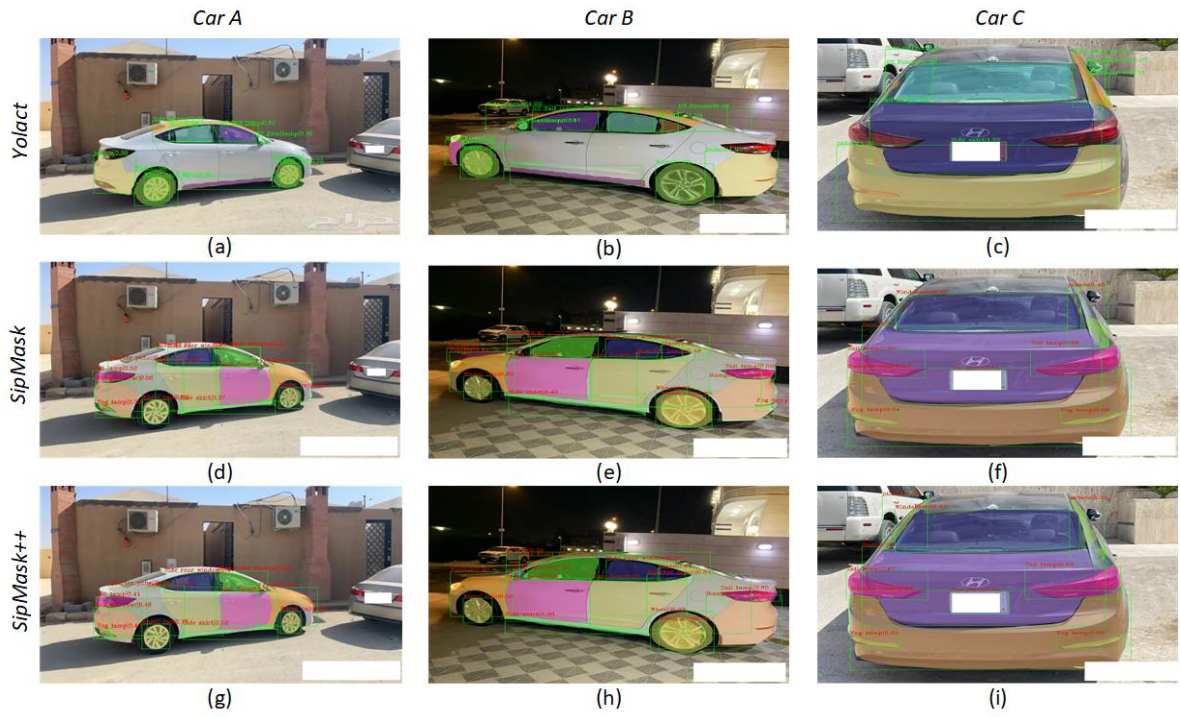

*Figure 9: Comparing Yolact to SipMask/SipMask++ instance segmentation regimes on a merged-classes (22) dataset with an increased sample size and augmentation modules. (a-c)Yolact, (d-f)SipMask, (g-i) SipMask++*

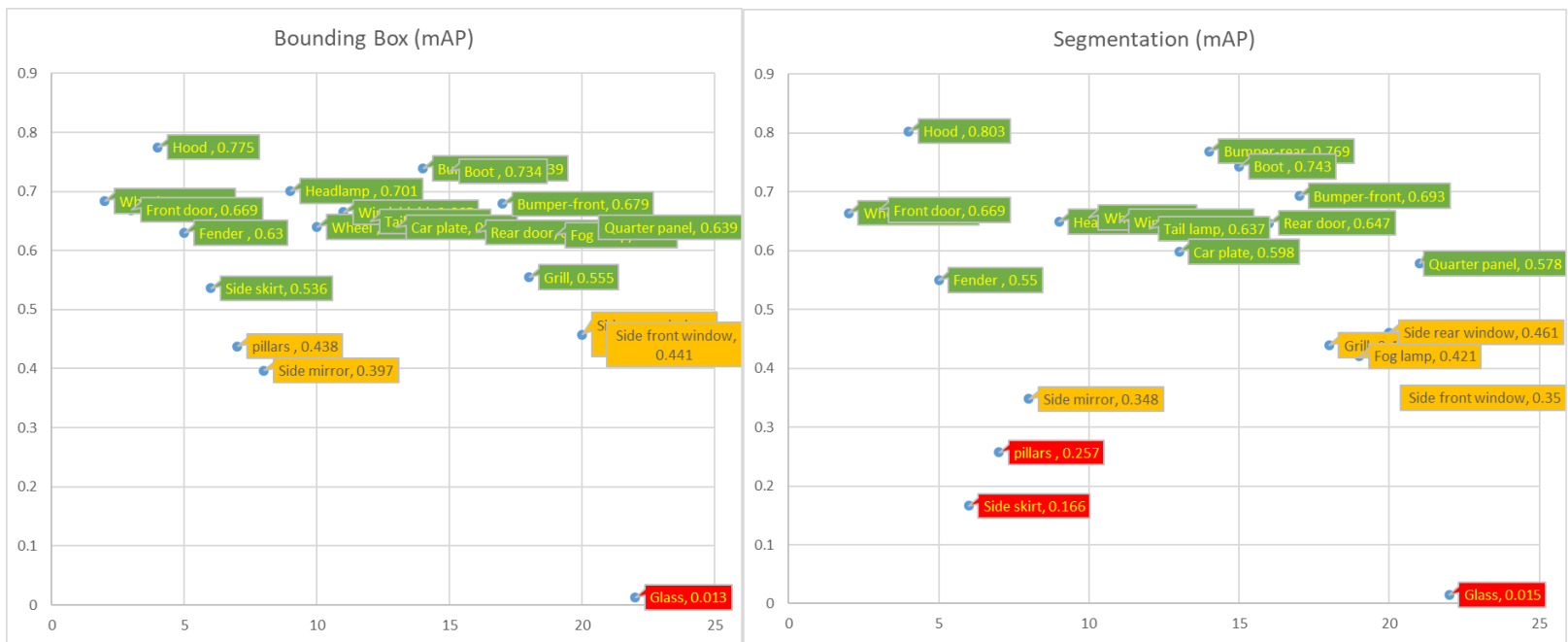

*Figure 10: Class-level mAP values for the Yolact algorithm*

### 3.6.4. Class-level performance for Yolact

Further looking into class-level mAP for the Yolact algorithm showed a marked performance loss in classes that were either very diverse (e.g., the Glass class due to its varied shapes and sizes as well its reflectivity) or had smaller cross-sections such as the Pillar or Side Skirt class (Figure 10). These three classes had mAPs of 20.3, 21.9, and 27.6 respectively compared to classes that had a higher likelihood of being similar for different models such as the Hood, Bumper-rear and Boot classes with mAPs (0.775, 0.739, and 0.734). The Glass class was indeed under-represented in the dataset but the remaining two classes, regardless of their abundance in the dataset, their masks were outliers to the remaining parts compared to their bounding boxes, which lead to lower small-object mAPs. In general, the larger and spatially similar objects showed a better mAP compared to smaller or thinner parts. The same observation held true for both detection and segmentation outcomes and hence the analysis was further extended to three randomly selected examples (Car A, Car B and Car C) shown in Figure 9 with the detailed mAPs shown in Table 4. The table shows valid mAPs as green cells with any misclassified cells in a pair (e.g. Wheel cap) shown as amber. Any undetected or incorrectly classified classes were shown as Red cells. A further precision/recall analysis of the three cars against the three algorithms further shows a marked difference between Yolact and SipMask genres with the latter category clearly outperforming for bounding box scales (Table 5).

|  | Bounding Box | | | Bounding Box | | | Bounding Box | | |
|---|---|---|---|---|---|---|---|---|---|
|  | Car A | | | Car B | | | Car C | | |
|  | Yolact | SipMask | SipMask++ | Yolact | SipMask | SipMask++ | Yolact | SipMask | SipMask++ |
| Wheel cap (P) | 0.87 | 0.6 | 0.69 | 0.5 | 1 | 0.58 | NV | NV | NV |
| Front door | FN | 0.61 | 0.52 | FN | 0.62 | 0.58 | NV | NV | NV |
| Hood | NV | NV | NV | FN | 0.31 | FN | NV | NV | NV |
| Fender | FN | 0.5 | 0.39 | FN | 0.41 | 0.35 | NV | NV | NV |
| Side skirt | Hood | 0.37 | 0.52 | Hood | 0.4 | 0.36 | NV | NV | NV |
| pillars | FN | 0.32 | FN | FN | FN | FN | NV | NV | NV |
| Side mirror | Headlamp | FN | FN | Headlamp | FN | 0.31 | NV | NV | NV |
| Headlamp | NV | NV | NV | NV | NV | NV | NV | NV | NV |
| Wheel (P) | 1 | FN | FN | 1 | 0.35 | 0.35 | NV | NV | NV |
| Windshield | NV | NV | NV | NV | NV | NV | Fender | 0.55 | 0.62 |
| Tail lamp | FN | 0.52 | 0.41 | FN | 0.6 | 0.6 | ND | 0.54 | 0.5 |
| Car plate | NV | NV | NV | NV | NV | NV | Side-skirt | 0.6 | 0.56 |
| Bumper-rear | Pillars | 0.56 | 0.48 | Pillars | 0.37 | 0.57 | Pillars | 0.63 | 0.69 |
| Boot | NV | NV | NV | NV | NV | NV | FN | 0.66 | 0.65 |
| Rear door | FN | 0.54 | 0.61 | FN | 0.72 | 0.63 | NV | NV | NV |
| Bumper-front | NV | NV | NV | NV | NV | NV | NV | NV | NV |
| Grill | NV | NV | NV | NV | NV | NV | NV | NV | NV |
| Fog lamp | FN | 0.39 | 0.42 | NV | NV | NV |  |  |  |
| Side rear window | Boot | 0.49 | 0.47 | Boot | 0.53 | 0.56 | NV | NV | NV |
| Quarter panel | FN | 0.43 | 0.52 | Car plate | 0.48 | 0.57 | NV | NV | NV |

| | | | | | | | | | | |
|---|---|---|---|---|---|---|---|---|---|---|
| Glass | FN | FN | FN | FN | 0.47 | 0.41 | | NV | NV | NV |
| Side front window | Tail-lamp | 0.42 | 0.45 | Tail-lamp | 0.45 | 0.36 | | NV | NV | NV |
| Precision | 3/(3+5) | 12/13 | 13/13 | 3/3+5 | 13/15 | 14/15 | | 1/3 | 5/5 | 5/5 |
| Recall | 3/12 | 12/17 | 13/17 | 3/11 | 13/17 | 14/17 | | 1/3 | 5/5 | 5/5 |

| Legend | |
|---|---|
| [score] | Part identified correctly |
| [av-score] | One part identified |
| NV | Not visible |
| ND | Not detected (False negative) |
| [Part-name] | Incorrectly detected (False positive) |

*Table 4: A part-level bounding box analysis of Dataset-3 performance*

| | Car A | | | Car B | | | Car C | | |
|---|---|---|---|---|---|---|---|---|---|
| | Yolact | SipMask | SipMask++ | Yolact | SipMask | SipMask++ | Yolact | SipMask | SipMask++ |
| Precision | 0.38 | 0.92 | 1.00 | 0.38 | 0.87 | 0.93 | 0.33 | 1.00 | 1.00 |
| Recall | 0.25 | 0.71 | 0.76 | 0.27 | 0.76 | 0.82 | 0.33 | 1.00 | 1.00 |

*Table 5: A precision/recall analysis of part-level accuracies for Cars A, B and C for the three algorithms including Yolact, SipMask, SipMask++*

## 4. Conclusion and future work

This research reports on the evaluation of three car parts datasets, one curated from publicly available images and the other organized as part of this research that was later extended and evaluated against a modified class distribution where classes with higher similarity were merged to minimize cross-class bias. The underlying algorithms used were selected based on their reported performance on another commonly benchmarked dataset. Two unique architectures, Yolact and SipMask were evaluated, and the selection was made also on the fact that Yolact originated from a two-stage genre of real-time instance segmentation mechanisms whereas the latter (SipMask) comprised of a single-stage instance segmentation approach that reported promising results on the COCO dataset for both bounding box and masking accuracies.

The initial results reported on Yolact presented the most balanced trade-off between performance and accuracy. There were certain cases of localization failures where each object's boundaries had inaccuracies particularly where the objects were similar and had thin segmentation lines. One such example was that of front and rear doors. There were also cases of mask leakage where the network did not manage to suppress noise outside the boundary box especially when two similar objects were away from each other. One such example was that of front and rear wheels. Once the classes were merged in a bid to minimize cross-class similarities and hence, unnecessary bias, the overall precision improve substantially.

The future direction of this work now aims to extend to the damage localization and assessment stage with the main objective focussing on modifying the semantic segmentation architectures to generate a combined classification and regression outputs for parts localization, damage localization and segmentation as well as parts and labour cost regression. Moreover, the research now further aims to focus on under-represented classes by the introduction of a synthetic dataset to allow the alteration of aspects such as colour, position, noise, and scale. The introduction of this synthetic dataset will also allow the extension of new car models to the existing training pipeline via transfer learning.